\documentclass[conference]{IEEEtran}
\IEEEoverridecommandlockouts
\usepackage{tikz}

\newcommand\copyrighttext{%
  \footnotesize \textcopyright \the\year{} IEEE. Personal use of this material is permitted. Permission from IEEE must be obtained for all other uses, including reprinting/republishing this material for advertising or promotional purposes, collecting new collected works for resale or redistribution to servers or lists, or reuse of any copyrighted component of this work in other works.}

\newcommand\copyrightnotice{%
\begin{tikzpicture}[remember picture,overlay]
\node[anchor=south,yshift=10pt] at (current page.south) {\fbox{\parbox{\dimexpr0.75\textwidth-\fboxsep-\fboxrule\relax}{\copyrighttext}}};
\end{tikzpicture}%
}

\usepackage{cite}
\usepackage{amsmath,amssymb,amsfonts}
\usepackage{array}
\usepackage{algorithmic}
\usepackage{graphicx}
\usepackage{textcomp}
\usepackage{xcolor}
\usepackage{listings}
\usepackage{verbatim}
\usepackage{balance}
\usepackage[most]{tcolorbox}
\usepackage{hyperref}
\usepackage[utf8]{inputenc} 
\usepackage{csquotes}
\MakeOuterQuote{"}
\usepackage[T1]{fontenc} 
\usepackage{mathpazo} 
\usepackage{pdflscape}
\usepackage{afterpage}
\usepackage{graphicx} 
\usepackage{booktabs}
\usepackage{multirow}
\usepackage{booktabs} 
\usepackage{tabularx}
\usepackage{makecell}
\usepackage{titlesec}

\usepackage{pdfpages}
\usepackage{listings} 
\usepackage{tcolorbox}
\usepackage{enumerate} 
\usepackage{transparent}
\usepackage{xurl} 
\usepackage{parskip}
\usepackage{float}
\usepackage{threeparttable}
\usepackage{tabularx}
\usepackage{vhistory}
\usepackage{hyperref}
\hypersetup{
    colorlinks,
    citecolor=black,
    filecolor=black,
    linkcolor=black,
    urlcolor=black
}
\usepackage{caption} 
\usepackage{amssymb}
\usepackage{pifont}

\def\BibTeX{{\rm B\kern-.05em{\sc i\kern-.025em b}\kern-.08em
    T\kern-.1667em\lower.7ex\hbox{E}\kern-.125emX}}
\begin{document}

\title{A Survey Study on the State of the Art of Programming Exercise Generation using Large Language Models}

\author{\IEEEauthorblockN{1\textsuperscript{st} Eduard Frankford}
\IEEEauthorblockA{\textit{Computer Science} \\
\textit{University of Innsbruck}\\
Innsbruck, Austria \\
eduard.frankford@uibk.ac.at}
\and
\IEEEauthorblockN{2\textsuperscript{nd} Ingo Höhn}
\IEEEauthorblockA{\textit{Computer Science} \\
\textit{University of Innsbruck}\\
Innsbruck, Austria \\
ingo.hoehn@student.uibk.ac.at}
\and
\IEEEauthorblockN{3\textsuperscript{rd} Clemens Sauerwein}
\IEEEauthorblockA{\textit{Computer Science} \\
\textit{University of Innsbruck}\\
Innsbruck, Austria \\
clemens.sauerwein@uibk.ac.at}
\and
\IEEEauthorblockN{4\textsuperscript{th} Ruth Breu}
\IEEEauthorblockA{\textit{Computer Science} \\
\textit{University of Innsbruck}\\
Innsbruck, Austria \\
ruth.breu@uibk.ac.at}
}

\maketitle
\renewcommand\fbox{\fcolorbox{red}{white}}
\setlength{\fboxrule}{2pt}
\copyrightnotice
\definecolor{main}{HTML}{808080}    
\definecolor{sub}{HTML}{d9d9d9} 
\tcbset{
	sharp corners,
	colback = white,
	before skip = 0.2cm,    
	after skip = 0.5cm      
}
\newtcolorbox{findingsBox}{
	colback = sub, 
	colframe = main, 
	boxrule = 0pt, 
	leftrule = 6pt 
}    

\newtcolorbox{promptBox}{
  colback = sub, 
  colframe = main, 
  boxrule = 1pt, 
  leftrule = 3pt,
}

\begin{abstract}

This paper analyzes Large Language Models (LLMs) with regard to their programming exercise generation capabilities. Through a survey study, we defined the state of the art, extracted their strengths and weaknesses and finally proposed an evaluation matrix, helping researchers and educators to decide which LLM is the best fitting for the programming exercise generation use case. We also found that multiple LLMs are capable of producing useful programming exercises. Nevertheless, there exist challenges like the ease with which LLMs might solve exercises generated by LLMs. This paper contributes to the ongoing discourse on the integration of LLMs in education.
\end{abstract}


\section{Introduction}

The advent of the mainstream use of LLM based applications such as ChatGPT has marked a significant milestone in the field of artificial intelligence. State of the art LLMs have demonstrated remarkable capabilities in a variety of applications, ranging from feedback creation to code generation \cite{frankford2024ai, bassner2024iris}. In the realm of programming education, the potential of LLMs to automate and enhance the generation of learning materials, such as programming exercises, is a promising area of exploration. Despite a significant number of publications, a dedicated survey study that examines and compiles these first attempts at generating programming exercises is still missing.

This research seeks to address this gap by analyzing and synthesizing state-of-the-art literature about the application of LLMs in the field of programming exercise generation. This study is therefore guided by the following research questions:

\begin{itemize}
\label{sec:RQs}
    \item \textbf{RQ1:} What is the current state of the art in programming exercise generation using LLMs?
    \item \textbf{RQ2:} What are the strengths and weaknesses of existing solutions in programming exercise generation using LLMs?
    \item \textbf{RQ3:} How can different LLMs be evaluated regarding their suitability for exercise generation?
\end{itemize}

The significance of this research lies in its potential to provide a more complete view of the topic of exercise generation, helping educators and researchers gain a better understanding of how to optimize educational resources in programming education. 

The structure of this paper is organized as follows: Section ~\ref{sec:relwork} offers an overview of selected related literature. Section ~\ref{sec:method} details the methodologies used. The core findings of the study are outlined in Section ~\ref{sec:results}, with a discussion of the results provided in Section ~\ref{sec:disc}. Key limitations of the study are addressed in Section ~\ref{sec:threats}, and the paper concludes with Section ~\ref{sec:conc}, encapsulating the primary insights and discussing the wider implications of the study.

\section{Related Work}
\label{sec:relwork}

The utilization of LLMs across various domains, including software engineering and education in general, has been extensively documented, with significant contributions from Hou et al. \cite{hou2023large}, Hadi et al. \cite{hadi2023survey}, and Yan et al. \cite{Yan_2023}. Hou et al. categorize LLMs' applications in software engineering, highlighting their adaptability, a feature crucial for educational applications like programming exercise generation. Hadi et al. extend this perspective by exploring LLMs' broader implications, challenges, and architectures, emphasizing their transformative potential and ethical considerations. Yan et al. conducted a comprehensive review on the ethical and practical challenges of LLMs in education, revealing their versatility across various educational tasks, from content generation to feedback provision.

While these studies collectively describe the capabilities, applications, and challenges of LLMs in software engineering and education, there is still no survey study specifically addressing the generation and benchmarking of programming exercises using LLMs. This research addresses this gap by exploring the potential of LLMs to automate programming exercise generation and by developing an evaluation matrix to guide educators in selecting the most appropriate LLM for their needs.

\section{Methodological Approach}
\label{sec:method}

This study made use of the methodological approach of a survey study to offer a complete view of the state of the art, highlighting significant contributions, and to identify potential areas for future research. The approach encompasses the following key components.

\subsection{Literature Collection Strategy}

To conduct a comprehensive search, both traditional academic databases and Google Scholar were utilized due to the novelty of the topic. The traditional databases included: (1) ACM Digital Library, (2) IEEE Xplore, (3) ScienceDirect, and (4) SpringerLink.
After searching the classical databases and Google Scholar, a round of snowballing was conducted \cite{Wohlin2014}.

\subsection{Search Terms}

The search terms were derived from the core topics of interest: LLMs and programming exercise generation. A combination of keywords related to "Large Language Models", "Programming Exercise Generation" and "Educational Technology" was used to capture the relevant literature. The search was iteratively refined to include emerging terms and synonyms to ensure comprehensiveness.

\subsection{Inclusion \& Exclusion Criteria}

In alignment with Garousi et al.'s \cite{Garousi2019} guidelines, this survey study adopts inclusion and exclusion criteria to ensure the relevance and timeliness of the literature reviewed. We focus on studies published from 2018 to 2023, reflecting the rapid developments in AI. We only selected studies that offer insights into LLM applications in programming exercise generation and are accessible in full text. Last but not least, only articles available in English were considered.

\subsection{Stopping Criteria}

The stopping criterion of data exhaustion was used. However, this survey study, examined the entire body of available literature for the defined search string. Given the limited scope of existing research, all relevant studies were covered before reaching a point of data saturation. This point is supported by the fact that even snowballing did not add new sources to the body of included literature.

\section{Results}
\label{sec:results}

This section presents the findings from the survey study, organized by research question.

\subsection{RQ1: Current State in Programming Exercise Generation}
\label{sec:result_current_state}

Working implementations for programming exercise generation exist and produce sensible results \cite{Sarsa2022, Freitas2023, Denny2022, haluptzok2023}. The study by Sarsa et al. \cite{Sarsa2022} offers the most detailed results, claiming LLMs are capable of generating sensible, novel, and readily applicable programming exercises. Of 120 investigated exercises, 75\% were sensible, 81.8\% were novel, and 76.7\% had a matching sample solution. Denny et al. \cite{Denny2022} reported similar results, which is expected since they used the same underlying model (Codex) and applied it to a similar task. Haluptzok et al. \cite{haluptzok2023} generated programming exercises, not with the intention of providing them to humans, but rather for model training purposes. The purpose was to enhance the code generation capabilities of the tested model through the creation and solving of programming exercises.

One reoccurring prompting technique is the decomposition of exercise parts into: (1) problem statement, (2) template code, (3) solution and (4) test cases. For exercise generation, this allows to reduce the context size and increases the likelihood of generating more precise results \cite{ZamfirescuPereiraLLMPrecision}. Freitas et al. \cite{Freitas2023} used two different models, for different components of the exercise. For the problem statement, the Google T5 model was used, while for the generation of template code, Google CodeT5 was employed. According to the researchers, this demonstrated equally capable results to OpenAI's GPT-3 model, while only using smaller open source models.

In general, programming exercise generation systems have been using various LLMs. Sarsa et al. and Denny et al. \cite{Sarsa2022, Denny2022} utilized Codex \cite{ChenCodex}, which is based on the GPT-3 \cite{brown2020} architecture but fine-tuned specifically for programming tasks, originally powering the GitHub Copilot. Freitas et al. \cite{Freitas2023}, in contrast, employed Google T5 and CodeT5 models. 
Haluptzok et al. \cite{haluptzok2023} experimented with the open-source GPT-Neo \cite{GPTneo} model in its various forms (125M, 1.3B, and 2.7B versions) along with Codex. GPT-Neo is similar to GPT-3 and was trained on the Pile dataset \cite{ThePile}, which includes a substantial amount of GitHub code, providing a robust foundation for programming-related tasks.

\subsection{RQ2: Strengths and Weaknesses of Existing Solutions}
\label{sec:result_strengths_weaknesses}

The main advantage of automated exercise generation lies in its remarkable ability to create learning material in a time efficient way. The traditional process of developing educational resources is labor-intensive and often demands a considerable degree of expertise. This efficiency opens up unprecedented possibilities for creating an extensive array of novel learning resources, including detailed code explanations and comprehensive code examples, on a virtually limitless scale \cite{Becker2023}.

Haluptzok et al. \cite{haluptzok2023} underscore the significant benefits of fine-tuning in the context of programming exercise generation. Specifically, fine-tuning Neo models on verified synthetic puzzle-solution pairs resulted in a two to five times improvement in puzzle-solving capabilities compared to the baseline model. This demonstrates a key strength of LLMs, a model with the general capacity of, for example GPT-4, when fine-tuned for programming exercise generation, could likely lead to impressive results.

Additionally, generated exercises, even if imperfect, can serve as a starting point for further modification by teachers or as a basis for student activities, like code reviewing and debugging exercises \cite{Sarsa2022}. Similar observations are made by Kasneci et al. and Prather et al. \cite{Kasneci2023, prather2023}. On the one hand, they emphasize personalizing the context of exercises, making them more engaging to the respective student. On the other hand, programming exercise generation enables more granular scaling by difficulty. 

A weakness mentioned by Sarsa et al. \cite{Sarsa2022} is the quality of the generated tests or test suites. Less than a third of exercises with tests pass successfully, and only about 70\% of the exercises include tests at all. Therefore, the general precision of the generation method has still to be improved.

Another significant challenge, mentioned by Denny et al. \cite{Denny2022}, is that exercises generated using LLMs seem to be easily solvable by LLMs as well. This aspect introduces the risk of creating a counterproductive cycle where the exercises produced by LLMs fail to adequately challenge students. This is because students might find it tempting to simply input these exercises into a LLM-based application, such as ChatGPT, and obtain solutions with minimal effort. 

Freitas et al. \cite{Freitas2023} mentioned a quality trade-off between costly commercial models like GPT-3, GPT-4 and freely accessible open source models, like Google T5 or GPT-Neo.

Other studies \cite{Becker2023, prather2023, Kasneci2023} highlight several disadvantages such as an over-reliance on the student as well as the teacher side, ethical concerns and licensing issues.

\subsection{RQ3: Evaluating LLMs Suitability for Exercise Generation}
\label{sec:result_benchmarks}

Most benchmarks for LLMs evaluate their performance based on a set of problems that the models are expected to solve. The evaluation is then conducted by calculating the percentage of correctly solved problems \cite{brown2020, hendrycks2021, ChenCodex, guo2023evaluating}. As the models are getting better, there has been the trend of changing benchmarks to contain more complex problems \cite{hendrycks2021, Wang2019}. Earlier benchmarks like the \textit{Winograd Schema Challenge} are being phased out since recent models neared human-level performance \cite{hendrycks2021}. Improved benchmarks, like the \textit{MMLU}, include harder and more specialized subjects \cite{hendrycks2021}.

In programming, the \textit{HumanEval} metric, introduced in the Codex paper by OpenAI researchers \cite{ChenCodex}, is a well-established benchmark \cite{guo2023evaluating}. A subsequent study comparing several LLMs' coding abilities also based its evaluation on \textit{HumanEval} \cite{Xu2022}. Liu et al. \cite{liu2023} proposed \textit{EvalPlus}, which builds upon \textit{HumanEval}, adding the ability to detect more incorrectly synthesized code.

During the survey study, we have found both manual and automated LLM assessment approaches \cite{Sarsa2022}. 
Regarding manual assessment, Sarsa et al. \cite{Sarsa2022} present four metrics: (1) Sensibleness, (2) Novelty, (3) Topicality and (4) Readiness for Use. \textit{Sensibleness} is defined as whether the problem could be given to students to solve. \textit{Novelty} is true if the programming exercise is not findable in Google or GitHub. \textit{Readiness of use} considered the amount of manual work needed by a teacher before being able to use the generated problem. The concept, \textit{topicality}, is related to \textit{novelty} and examines how well a concept given in the prompt is accounted for in the created exercise.

The studies by Sarsa et al. and Denny et al. \cite{Sarsa2022, Denny2022} also reported five metrics that can be evaluated automatically: (1) Has sample solution, (2) Sample solution is executable, (3) Has tests, (4) All tests pass and (5) Test coverage.

The general drawbacks of a manual evaluation are subjectivity, bias and cost when compared to automated evaluations \cite{ChenCodex}. 

Having analyzed multiple approaches for benchmarking in general, it is important to derive a benchmark tailored to benchmarking programming exercise generation instead of general programming abilities of LLMs.

As a first step, it is important to define the key requirements of programming exercise generation. Only then, metrics can be established to measure model performance.
Defining what constitutes a good exercise, which results in effective learning outcomes for students, is challenging. Typically, high-quality exercises share characteristics such as a well-defined problem statement, good structure, a specific competency focus and a progressively increasing difficulty for sub-tasks. 
Additionally, for the use in higher education, costs need to be taken into account, because decomposing the exercises and including previously generated exercise parts in the prompt, can lead to expensive requests when using paid models like GPT-4 \cite{Freitas2023}. 
The constructs of novelty and readiness of use have been proposed before and should also be taken into account to measure the quality of the generated exercises \cite{Sarsa2022, Denny2022}.  
Additionally, a viable approach to establish a quantitative benchmark for programming exercise generation using LLMs could be a match-based multiple choice approach, similar to the \textit{hellaSWAG} benchmark \cite{Zellers2019}. The \textit{hellaSWAG} dataset evaluates the ability to complete unfinished sentences with multiple choice questions. Similarly, a dataset could be developed consisting of partial exercises. The model would then need to determine which of the provided answers best completes the exercise. This approach could be implemented as an automated benchmark, and named the \textit{Programming Exercise Generation Benchmark (PEGB)}. 

A more complete picture of LLM performance for programming exercise generation can be formed when evaluating a sample of generated exercises using the evaluation matrix presented in Table \ref{tab:matrix}. This takes into account general dimensions like cost, ability to generate code and data privacy, as well as specialized benchmarks like the before defined \textit{PEGB} and constructs like sensibility, novelty and readiness of use. Using this matrix, educators may form a more evidence-based choice for a LLM to power their programming exercise generation service. 

\begingroup
\begin{table}[]
\small
\centering
\begin{threeparttable}
\caption{Evaluation Matrix}
\label{tab:matrix}
\begin{tabular}{|l|l|}
\hline
& \textbf{LLM} \\ \specialrule{.2em}{.1em}{.1em}
\textbf{General Assessment} &   \\ \specialrule{.2em}{.1em}{.1em}
Costs / Semester    &     Numeric               \\ \hline
Data Privacy   &     Boolean             \\ \hline
HumanEval Score  &   Percentage                \\ \hline
PEG Benchmark        &         Percentage               \\ \specialrule{.2em}{.1em}{.1em}
\textbf{Program Analysis}        &         \\ \specialrule{.2em}{.1em}{.1em}
Has sample solution        &        Percentage              \\ \hline
Runnable sample solution       &         Percentage               \\ \hline
Test Cases are present       &        Percentage             \\ \hline
All tests pass       &         Percentage               \\ \hline
Test coverage       &        Percentage               \\ \specialrule{.2em}{.1em}{.1em}
\textbf{Qualitative Assessment} & \\ \specialrule{.2em}{.1em}{.1em}
Sensibility      &       Percentage                \\ \hline
Novelty         &      Percentage                \\ \hline
Readiness of Use &       Percentage                 \\ \hline
\end{tabular}
\end{threeparttable}
\end{table}
\endgroup

\section{Discussion}
\label{sec:disc}

The results presented in Section \ref{sec:results} provide a starting point to better understand the current landscape and potential of generative AI in programming education, particularly in the context of programming exercise generation. 

The current state of the art in programming exercise generation (RQ1), with Sarsa et al. \cite{Sarsa2022}, Freitas et al. \cite{Freitas2023}, Denny et al. \cite{Denny2022}, and Haluptzok et al. \cite{haluptzok2023}, reveal a promising trend towards the practical application of LLMs to improve educational resource creation. These studies demonstrate the capability of LLMs to produce novel, sensible, and ready-to-use programming exercises, which could significantly reduce the workload on educators and enable personalized learning experiences for students, which can be identified as their main strength. 
However, there are also significant weaknesses mentioned in the literature, such as the large failure quote of generated test suites and the potential for generated exercises to be easily solvable by LLMs themselves.

Last but not least, we found that there exist both automated and manual assessment approaches, which reflect the complexity of assessing AI-generated content's educational value. 

\section{Limitations}
\label{sec:threats}

This study's comprehensiveness might be limited by the search strategy and terms used to gather relevant publications. Although a broad range of keywords and search techniques, including snowballing, were employed to ensure an extensive collection, there's still a possibility that some important studies were overlooked. Notably, the snowballing process did not reveal any new documents, suggesting the initial search was thorough.

Additionally, the initial screening and selection of studies were primarily conducted by a single individual, raising concerns about potential subjective bias in choosing which studies to include. This risk was partially mitigated through feedback from other researchers, yet the potential for bias cannot be fully discounted.

\section{Conclusion}
\label{sec:conc}

This study explored the use of LLMs for generating programming exercises, highlighting their potential to transform programming education. LLMs such as Codex and GPT-3 can produce engaging and novel exercises. However, they often require manual refinement to be classroom-ready. Despite this, using LLMs for exercise generation offers a substantial time saving possibility for educators and provides the flexibility to customize exercises to effectively meet student needs.

Nevertheless, over-reliance on LLMs poses risks to exercise quality and student learning outcomes, particularly as it seems like LLM-generated exercises are easily solvable by other LLMs.

To navigate the evolving landscape of LLMs in education, we proposed an evaluation matrix to provide a structured approach for assessing LLMs regarding their exercise generation capabilities. This matrix considers factors such as cost, data privacy, code generation capabilities, and the newly introduced concept of a programming exercise generation benchmark. 

Looking ahead, completing the PEGB multiple choice programming exercise test set and regularly using the evaluation matrix to assess emerging LLMs will be crucial to help educators and researchers to identify the most effective models for their needs.

\bibliographystyle{IEEEtran}
\balance
{\small\bibliography{bibliography}}

\end{document}